# Medicine Strip Identification using 2-D Cepstral Feature Extraction and Multiclass Classification Methods

Anirudh Itagi, Ritam Sil, Saurav Mohapatra, Subham Rout,
Bharath K P, Karthik R, Rajesh Kumar Muthu

School of Electronics Engineering, Vellore Institute of Technology, Vellore.
anirudhitagi3@gmail.com, ritam.sil101@gmail.com, sauravmohapatra.2015@vit.ac.in, subham.rout2015@vit.ac.in,
bharathkp25@gmail.com, tkgravikarthik@gmail.com, mrajeshkumar@vit.ac.in

*Abstract*— **Misclassification of medicine is perilous to the health of a patient, more so if the said patient is visually impaired or simply did not recognize the color, shape or type of medicine strip. This paper proposes a method for identification of medicine strips by 2-D cepstral analysis of their images followed by performing classification that has been done using the K-Nearest Neighbor (KNN), Support Vector Machine (SVM) and Logistic Regression (LR) Classifiers. The 2-D cepstral features extracted are extremely distinct to a medicine strip and consequently make identifying them exceptionally accurate. This paper also proposes the Color Gradient and Pill shape Feature (CGPF) extraction procedure and discusses the Binary Robust Invariant Scalable Keypoints (BRISK) algorithm as well. The mentioned algorithms were implemented and their identification results have been compared.**

## I. INTRODUCTION

The irreplaceable sense of vision is vital for one to be aware of the surroundings their surroundings as well as to interact with it. Many people have some degree of blindness or poor vision. According to a review conducted by Bourne et al [1] about 36 million people are blind and 216.6 million have moderate to severe visual impairment. Many day to day tasks can only be performed correctly with eyesight such as reading, searching or watching activities. One such imperative task is to identify medicine strips and consume the right one at the right time. Ingestion of the wrong medicine can lead to the following problems:
1. The drug can counteract another previously taken medicine.
2. Upset the prescription cycle.
3. It may worsen the medical condition.
4. In severe cases, cause death due to negative drug reaction and poisoning.

Thus, efforts have been put into ensuring that medicine identification is progressing. The field of image processing extending into image classification and identification is a promising avenue for medicine strip, medicine box and medicine pills and capsules identification.

Machine learning and computer vision are powerful techniques that help in image interpretation, analysis and classification. The disadvantage with these techniques is their real time implementation. This is due to the execution time for the algorithm and also because of the massive computation power it draws. Computational effort needs to be optimized by improving the algorithms and exploring better processors.

Recent research works in image identification have been done by extraction - Scale Invariant Feature Transform (SIFT) and Speed Up Robust Features (SURF) features. SIFT is a computer vision algorithm that detects and describes features extracted as a large collection of feature vectors [2]. The feature vectors, if precisely extracted, are detectable even under image enhancement in terms of intensity of illumination, addition/removal of noise, or scaling of the image [3-4].

Benjamin et al [5] developed a system that uses SIFT as well as SURF based on Hessian Detector to extract the features. To match the features, their Haar wavelet responses are computed within the feature scale and the feature matching phase is performed using k-decision trees using the FLANN library.

Post feature extraction, feature matching is done to identify an image by comparing with a database of features extracted from a large set of images.

Horak et al [3] employed several multiclass classifying algorithms like the SVM, Decision tree, and Naive Bayes classifier and so on to evaluate the "interesting points" that were extracted by SURF method. They achieved a prediction accuracy of 98.15% using SVM and 99.80 % using the Naive Bayes classifier. Loussaif et al [4] have also applied SURF extracted features to SVM and variants of the SVM kernel as well as K-NN classifier demonstrating the excellent pairing that feature extraction and machine learning have and the applications they can lead to. SVM with cubic kernel gave an accuracy of 90% and weighted K-NN gave an accuracy of 67.5%.

This paper proposes two feature algorithms that perform medicine strip identification by means of: the first procedure being 2-D cepstral feature extraction coupled with machine learning and a second procedure, Color Gradient and Histogram of Oriented Gradients (HOG) features coupled with machine learning. This paper further builds on the blend of feature extraction and machine learning. Three feature extraction procedures have been employed, two procedures' extracted features are fed to K-NN, SVM and multiclass logistic regression classifiers and their outputs are tabulated. The third technique is that of the BRISK method, it involves BruteForce and BruteHamming to match features and feature descriptors.

The rest of the paper is organized as follows: Section II discuss the machine learning aspect of this paper where the extracted features are fed to the classification algorithms, section III discusses the BRISK algorithm, section IV is about the proposed CGPF extraction, section V is about the proposed 2-D cepstral feature extraction and the results are discussed in section VI, finally section VII concludes the paper.

## II. CLASSIFICATION METHODS

Most accurate image identification algorithms are obtained using machine learning. Supervised machine learning algorithms are employed for classification and pattern recognition tasks. Once features have been extracted using either of the algorithms that are discussed in section 4 and section 5, their features are stored for training and testing purpose of the classifier. The training to testing split here was 80% and 20% respectively. The classification methods used in this paper are discussed below:

### A. K-Nearest Neighbour

KNN is a popular instance-based, non-parametric learning method primarily used for classification [12]. It classifies an object by a majority vote among its neighbors and identifies the object as one among the neighbors' class. The distance metric is that of Euclidean distance and it is quite effective here owing to the multidimensional vector space of the features extracted.

The multidimensional distance is the parameter used to identify the class commonly the Euclidean distance is measured using:

$$d(x,y) = \sum_{i=i}^{k} \sqrt{x_i^2 + y_i^2} \qquad (1)$$

### B. Support Vector Machine

SVM is type of linear classifier [13], a class of supervised learning model, which is used to separate the data into different classes by a hyper plane for N dimensional data. SVM can perform nonlinear classification using kernel trick, thereby mapping the input features to higher dimensional feature spaces. SVM classification is done by cost function (J) minimization that indicates the most likely class the features belong to:

$$J(\theta) = -\frac{1}{m}\left(\sum_{i=1}^{m}\max(0,1 - y_i(\theta x_i - b))\right) + \lambda\|\theta\|^2 \qquad (2)$$

Where, y is the target class, x the corresponding features, X the matrix of features, θ the weight matrix and, b a bias constant, λ a regularization parameter that prevents overfitting of the cost function.

### C. Logistics Regression

Logistic regression or multinomial logistic regression generalizes to multiclass classification by apprehending a vector of variables and evaluating the coefficients or weights [14] for each input feature that has been extracted previously and then identifies the medicine. LR is done by cost function (J) minimization that indicates the most likely class the features belong to:

$$J(\theta) = -\frac{1}{m}\left(\sum_{i=1}^{m} y_i \log(h(x_i)) + (1 - y_i)\log(1 - h(x_i))\right) \qquad (3)$$

Where, y is the target class, x the corresponding features, X the matrix of features, θ the weight matrix and,

$$h = sigmoid(\theta * X) \qquad (4)$$

Each of the mentioned classification models are trained and tested separately with features extracted from the 2-D cepstral approach and separately with the features from the color gradient and HOG features method.

## III. BRISK IMAGE CLASSIFICATION

BRISK produces binary feature vectors and to compare the descriptor, Hamming distance is used. A set of key points and descriptors is extracted from the test image. These key points and descriptor are then further matched against database to find the best match. A Feature, here, is basically a place in image that can be easily (or) correctly recognize if another image of the same outlook is given. The features or the key points are taken from an image where there are inimitabilities in the image. In the case of medicine strips the features are not just the colors, but any unique points on the strips, letters or markings on the strips or the shape of pill itself.

RANSAC (Random Sample Consensus) algorithm is used to randomly sample from a set of potential matches and validate the similarity of the key points and doubly ensure if the key points are matched or not.

## IV. PROPOSED COLOR GRADIENT AND PILL SHAPE FEATURE EXTRACTION

The Color Gradient and Pill shape feature (CGPF) extraction procedure discussed in this paper is a two step process; it first involves obtaining the color gradient data of the image of the medicine strip and pill shape feature extraction using the Histogram of Oriented Gradients (HOG) method.

### A. Color Histogram

Color histogram is a technique generally used in image processing and photography to find the distribution of color pixels in an image [8]. It can be made for any kind of color image including gray-scale but three dimensional spaces like RGB or HSV. For monochromatic images it is generally known as intensity histogram [9]. In this paper color histogram technique is used to identify the color of the sample medicine strips and also the medicines if visible. This will be done after the HOG feature retrieval which will be used identify parameters like shape, area, perimeter, etc. Instead of plotting a histogram the peak value or maximum number of pixel at a particular shade of a color is measured and fitted onto a line graph.

$$r = \frac{n\sum uv + \sum u \sum v}{\sqrt{(n\sum u^2 - (\sum u)^2)(n\sum v^2 - (\sum v)^2)}} \quad (4)$$

r = correlation among the data

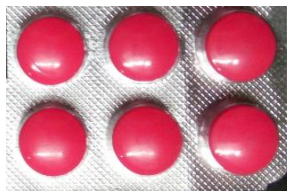

Fig. 1(a)

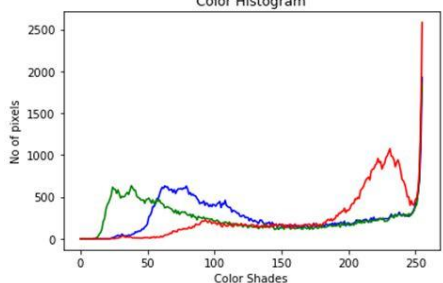

Fig. 1(b)

Fig.1. (a) Medicine strip imager, (b) the color analysis of this image.

### B. Histogram of Gradients

HOG is method which counts occurrences of gradient orientation of the contents inside an image. This method is very widely used in the fields of computer vision and image processing. This method has been applied over a test image which computes the features that represent the magnitude and orientation of the pixels in the image. Let $f(x, y)$ denote color intensity each of the pixel points of an image, then by applying HOG over the image the magnitude is obtained, $|\nabla f(x, y)|$ and orientation angle $\Theta(x, y)$ of the gradient, $\nabla f(x, y)$ [10].

The fundamental idea is that local object appearance and shape can often be identified by the distribution of local edge directions and/or gradient intensities. This is executed by dividing the image window into small spatial regions (or cells), each cell collects a narrow 1-D histogram of direction of gradients or edge orientations over the pixels of the cell.

The combined histogram entries structure the visualization in order to obtain better indifference to illumination, shadowing, etc., it is also useful to contrast-normalize the local responses before applying them. This can be done by accumulating a measure of localized histogram energy over a fairly larger spatial region (or block) and using the results to normalize all of the cells in the block. The normalized descriptor blocks will be referred to as HOG descriptors [11].

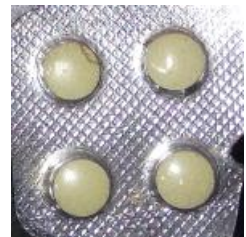

Fig. 2(a)

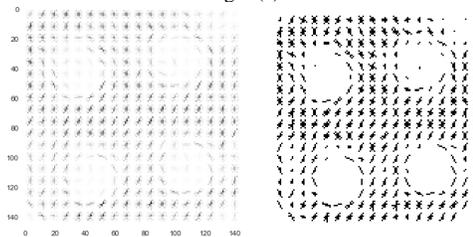

Fig. 2(c)  Fig. 2(b)

Fig. 2.(a) Original Image of Medicine Strip, (b) HOG Visualization of the medicine strip based on color gradients and (c) binary HOG visualization

The HOG visualization was used for feature extraction. First a median filter was used to remove the noise present. Then an initial thresholding has been done to remove all small speckles of dots.

Image Segmentation is a technique used to divide the image into a set of regions (tablets/pills inside the medicine strips) that are useful for the image identification. Segmentation can be of different types– Threshold based Segmentation, Region based Segmentation, Cluster based segmentation and Matching based segmentation.

Thresholding techniques recognize regions having similar pixel intensities which help to provide boundary to separate region of image having contrast background. Thresholding gives a binary output image from a given grayscale image (in this case the HOG visualization).

In this paper, Global Thresholding technique has been used to distinguish the HOG visualization image based on light objects and dark background. The pixel intensities of light objects and dark background are set as global threshold level, the image is masked to a binary image as follows:

$$g(x, y) = \begin{cases} 0, & f(x, y) < T \\ 1, & f(x, y) \geq T \end{cases} \quad (5)$$

Where, g(x, y) = pixel value of binary image obtained after thresholding at x and y, f(x, y) = pixel value of original image at x and y coordinates.

Finally, to obtain the accurate pill shape dimensions, RegionProps tool box is used, which is an image processing toolbox in MatLab used to measure properties of specific regions inside an image. The binary images are fed to the toolbox to mask out the light intensity regions and measure the geometrical parameters like area, perimeter, axis lengths, eccentricity etc. of those masked out regions. These output parameters are considered to be features (representing geometrical characteristics of required portions of image).
Figure 3 illustrates the CGPF algorithm.

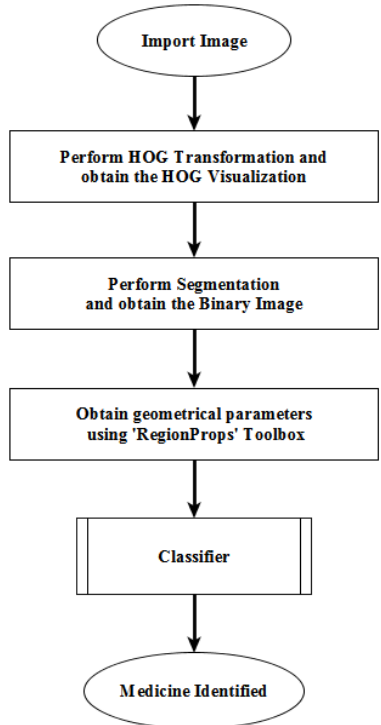

Fig. 3. CGPF feature extraction algorithm flow diagram

## V. PROPOSED 2-D CEPSTRUM FEATURE EXTRACTION

A database of images of several medicine strips has been made. The images compiled here were of different shapes and sizes, and of varying quality. The feature extraction procedure for and image is as follows: the selected image is split into its corresponding RBG (Red Blue Green) planes and the pixel values are taken as 2-dimensional (2-D) array. That is, the 3-D array, the image, is split into the 3 2-D arrays of pixel values. Each of these arrays are subject to a two dimensional Discrete Time Fourier Transform, using the 2-D Fast Fourier Transform. The absolute magnitude value of each pixel is taken and they undergo logarithmic compression. Because more signal energy is present at lower frequencies than higher frequencies. The resulting arrays are put together and from the resulting image bin values are obtained.

Bins are constructed comprising of discrete color pixel values. These bin values contain all valuable information regarding the image. The unique features lying within the obtained bin values of the image are distinct to it. In order to extract the vital features among them, the Discrete Cosine Transform (DCT) is applied. The first 20 coefficients of the resultant of this transform give the dominant features of the image that make the image unique. These features account for colors and the tonal distribution of the colors as well as the texture by perceiving the undetected contrasts the may exist in the image.

The feature extraction is comprised of the following steps:
1. 2-D Fast Fourier Transform to the RBG matrices
2. Logarithmic Compression of the magnitude of the values obtained from previous step
3. Obtain the Bin values of uniform sizes.
4. Take the Discrete Cosine Transform of the Bin values and extract the first 20 coefficients. These are the predominant features of the images.

The following flow diagram illustrates the proposed 2-D cepstral algorithm:

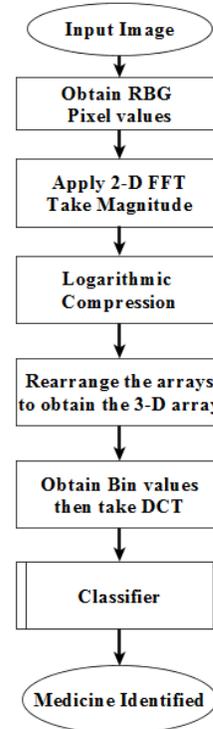

Fig. 4. The proposed 2-D cepstral algorithm

### A. The 2-D Fast Fourier Transform

2-D FFT is very popular for digital processing. It is frequently applied to obtain the spatial frequency of images, image power spectrum [6]. The 2-D FFT is characterized by the following equation:

$$F(k,l) = \frac{1}{MN}\sum_{x=0}^{M-1}\sum_{y=0}^{N-1} f(x,y) e^{-i2\pi\left(\frac{kx}{M}+\frac{ly}{N}\right)}. \quad (6)$$

Where, f(x,y) is the pixel matrix the transformation is applied to. M×N are the dimensions of the pixel array. F(k,l) is matrix obtained after transformation. Figure 5 shows a medicine strip and the logarithm of the 2-D FFT of the image.

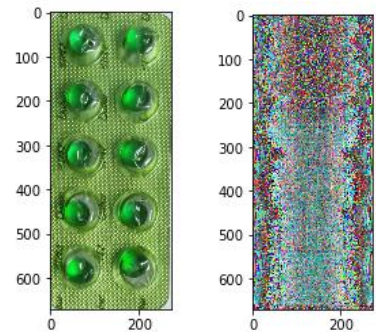

Fig. 5.(a)        Fig. 5.(b)

Fig. 5: (a) Image of the medicine strip (b) image after applying 2-D FFT and logarithmic compression

### B. Bin Values

After obtaining the image spectral density data vis-à-vis the 2-D FFT of the image, it is categorized into bins of uniform width and the numbers of bins are taken into consideration.

These bin values' data can correspond to the color gradients, texture representation and other decisive features of the image, if extracted precisely. Figure 3 shows the bin data of a set of 12 images.

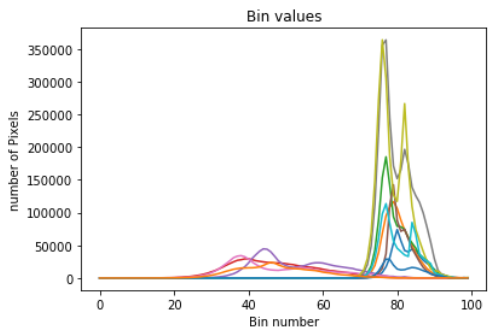

Fig. 6. Comparison of Bin values of a set of 12 medicine strips

## C. Discrete Cosine Transform

The DCT of a function can be defined as the sum of cosine functions oscillating at various frequencies. Its applications range from spectral analysis of images and audio files to obtaining numerical solutions of partial differential equations [7]. Unlike the Fourier transform, the DCT gives and relates using only real numbers. Here, the resultant of the DCT gives those features, which define the image and make it what it is. Of all the features extracted, due to the nature of the DCT, the first 20 are dominant enough to clearly identify the image. Figure 7 illustrates the DCT of a set of 12 medicine strips.

$$F(u) = \left(\frac{2}{N}\right)^{0.5} * \sum_{i=0}^{N-1} \Lambda(i) \cos\left(\frac{\pi u}{2N}(2i+1)\right) f(i) \quad (7)$$

Where, $f(i)$ is the function the DCT is applied to, $F(u)$ is the function obtained after the transformation and $\Lambda(i)$ is a thresholding function defined by:

$$\Lambda(i) = \begin{cases} \frac{1}{\sqrt{2}} & \text{if } i = 0 \\ 1 & \text{if } i \neq 0 \end{cases} \quad (8)$$

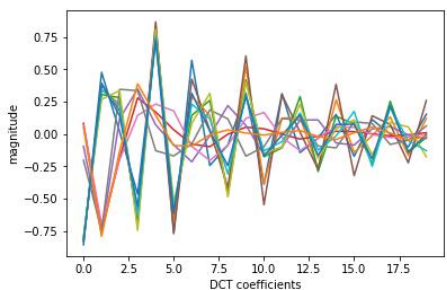

Fig. 7. Comparison of the DCT of the Bin values of a set of 12 medicine strips

## VI. RESULTS AND DISCUSSIONS

Post feature extraction, the obtained 2-D cepstral coefficients are stored, compiled and a database is made. And similarly, another database for the CGPF features is also made. The databases created for this paper contains the coefficients of over 520 medicine strips' images. The coefficients are fed to the classifier which identifies the medicine. Figure 8 shows a small sample of the images. After their feature extraction, 80% of the data was set for training the classifiers and 20% was for testing them. The classifiers used were the K-NN, Support Vector Machine (Linear Kernel) and Logistic Regression multiclass classifiers. Table 1 shows the accuracies of correctly identifying the medicine by looking at the medicine strip. Table 2 compares that of all three algorithms.

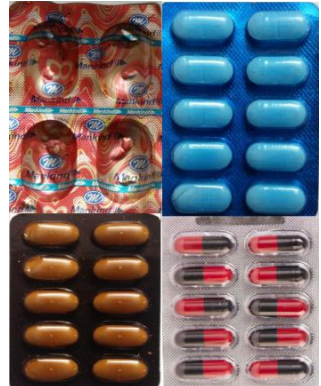

Fig. 8: Images Dataset sample

Table 1: Comparison of proposed 2-D Cepstrum and CGPF algorithms accuracy of identification

| Classifier | 2-D cepstrum | CGPF |
|---|---|---|
| K-NN | 99.82% | 95.65% |
| SVM | 99.734% | 94.56% |
| LR | 89.786% | 82.61% |

Table 2: Accuracy of Medicine Identification

| Algorithm | Classification Accuracy |
|---|---|
| 2-D cepstral | 99.82% |
| CGPF | 95.56% |
| BRISK | 98.86% |

To test the response of the response of the algorithm with different data sizes, the number of medicine strip image samples were varied and fed to the classifiers, the response is as shown in Figure 9. Clearly, if the data size goes less than 30, no classifier can identify what the image is, because the data size is comparable to the number of parameters resulting in insufficient training of the classifier, thereby rendering it incapable of classifying any image.

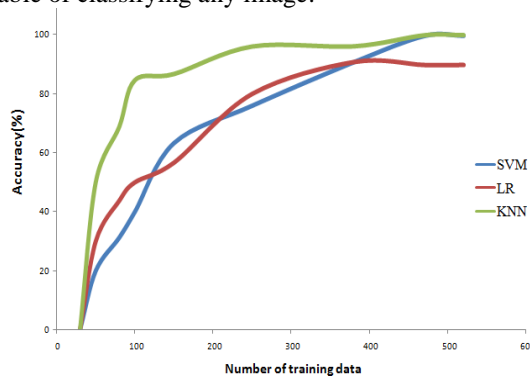

Fig. 9. Accuracy (in %) Vs. Data size

Furthermore to test the sensitivity of the algorithm, two remarkably similar looking medicine strips med05 (Oflochoice-OZ) and med08 (Oflokem-OZ) were taken. As shown in Figure 10, to even fully functioning naked eye, they look the same however on closer observation, med05 is more of a brownish-orange hue and med08 is of a brighter orange hue.

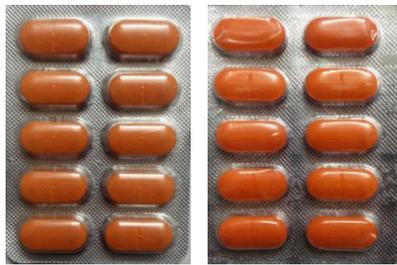

Fig. 10(a)            Fig. 10(b)
Fig. 10.(a) is med05 and (b) is med08

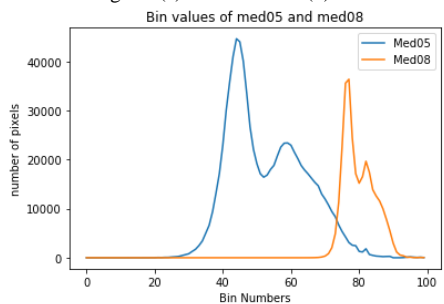

Fig. 11. Comparison of the bin values of med05 and med08

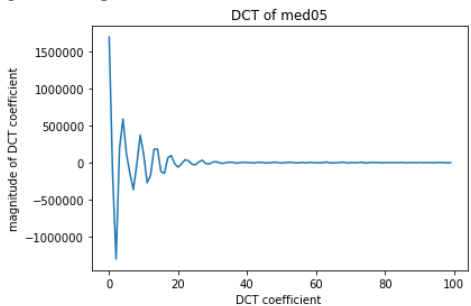

Fig. 12(a)

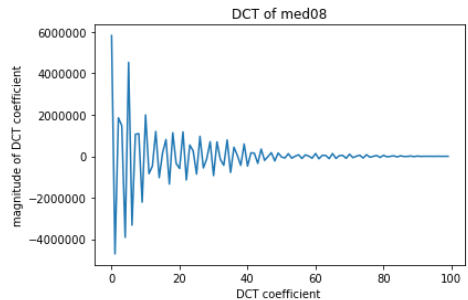

Fig. 12(b)
Fig. 12. Comparison of the DCT of med 05 and med08

But as seen from Figure 11 and Figure 12, clearly, the Bin values and the DCT of those bin values is paramount to the difference between the medicine strips.

It is because of unnoticed features like these that often medicines are mixed and the wrong medicine is ingested by the patient. This algorithm ensures that patient be safe from small mistakes like this that have large repercussions.

## VII. CONCLUSION

This work proposes two image classification algorithms and discusses a third procedure. The results of all three are exceedingly viable for real time implementation. It can also be concluded that the 2-D cepstral feature extraction procedure gave the best result, with an accuracy of 99.82% when coupled with the KNN classifier because of the nature of the features and the association it has with the image itself. It has also demonstrated the robustness and sensitivity it possesses with all sorts of medicine strips.